\documentclass{article}

\usepackage{arxiv}

\usepackage[utf8]{inputenc} 
\usepackage[T1]{fontenc}    
\usepackage{hyperref}       
\usepackage{url}            
\usepackage{booktabs}       
\usepackage{amsfonts}       
\usepackage{nicefrac}       
\usepackage{microtype}      
\usepackage{lipsum}
\usepackage{graphicx}
\graphicspath{ {./images/} }

\usepackage{times}
\usepackage{tablefootnote}
\usepackage{latexsym}

\usepackage{microtype}

\usepackage{xcolor}
\definecolor{mypink1}{rgb}{0.858, 0.188, 0.478}

\usepackage{subfigure}
\usepackage{color}
\definecolor{blue}{RGB}{0, 93, 170}
\usepackage{graphicx}
\usepackage{arydshln}
\usepackage{booktabs}
\usepackage{todonotes}
\usepackage{verbatim}
\usepackage{multirow}

\usepackage{enumitem}

\title{Neural Correction Model for Open-Domain Named Entity Recognition}

\author{
 Mengdi Zhu \\
  Peking University\\
  \texttt{1600012990@pku.edu.cn} \\
  \And
 Zheye Deng \\
  Peking University\\
  \texttt{dzy97@pku.edu.cn} \\
  \And
 Wenhan Xiong \\
  University of California, Santa Barbara\\
  \texttt{xwhan@cs.ucsb.edu} \\
  \And
  Mo Yu \\
  IBM Research \\
  \texttt{yum@us.ibm.com}\\
  \And
  Ming Zhang \\
  Peking University \\
  \texttt{mzhang\_cs@pku.edu.cn} \\
  \And
  William Yang Wang \\
  University of California, Santa Barbara \\
  \texttt{william@cs.ucsb.edu}
}

\begin{document}

\maketitle
\begin{abstract}
Named Entity Recognition (NER) plays an important role in a wide range of natural language processing tasks, such as relation extraction, question answering, etc. However, previous studies on NER are limited to particular genres, using small manually-annotated or large but low-quality datasets. Meanwhile, previous datasets for open-domain NER, built using distant supervision, suffer from low precision, recall and ratio of annotated tokens (RAT).
In this work, to address the low precision and recall problems, we first utilize DBpedia as the source of distant supervision to annotate abstracts from Wikipedia and design a neural correction model trained with a human-annotated NER dataset, DocRED, to correct the false entity labels. In this way, we build a large and high-quality dataset called AnchorNER and then train various models with it. To address the low RAT problem of previous datasets, we introduce a multi-task learning method to exploit the context information. We evaluate our methods on five NER datasets and our experimental results show that models trained with AnchorNER and our multi-task learning method obtain state-of-the-art performances in the open-domain setting.
\end{abstract}


\section{Introduction}
Named entity recognition (NER) aims to identify named entities, such as persons, locations, organizations, etc, from texts~\cite{DBLP:conf/coling/YadavB18}. As a key component of many natural language processing (NLP) tasks such as data mining, summarization, and information extraction \cite{DBLP:journals/computer/ChenCXWQC04,DBLP:conf/ijcai/BankoCSBE07,DBLP:conf/bionlp/AramakiMTOMO09}, NER has drawn much attention and many studies have been conducted in this field. In practice, NER could be applied to texts of various genres. Therefore, an NER model, especially an NER toolkit, should get high enough results in almost every case. However, open-domain is a great challenge for NER because large and high-quality NER datasets are very limited.

Previously, there are two main ways to address the problem. The first is using external knowledge and pre-training. Some researchers utilize knowledge bases, such as Freebase \cite{DBLP:conf/coling/GhaddarL18} and the Google Knowledge Graph \cite{DBLP:conf/lrec/BowdenWOMW18}. However, building up such resources is time-consuming and the coverage of a knowledge base is also limited. As for pre-trained embeddings, \cite{DBLP:conf/coling/MaCG16} presents a pre-trained label embedding method and a zero-shot framework that can predict both seen and previously unseen entity types.
Recently, pre-training language models \cite{DBLP:conf/naacl/DevlinCLT19} are also implemented in NER to significantly improve the performances. These strong models, nonetheless, only learn word representations without the information regarding named entities. After the pre-training process, we still need NER datasets to fine-tune the models in order to obtain better NER performances. 

The second way is increasing the training data with semi- or auto-labeled data. In \cite{DBLP:journals/ai/NothmanRRMC13}, authors build WikiNER using a semi-supervised method
and in \cite{DBLP:conf/ijcnlp/GhaddarL17}, researchers build an auto-labeled dataset called WiNER.
However, these datasets suffer from low qualities, including low precision, recall, and ratio of annotated tokens.

In this work
, our solution is along the direction of combining the best of both worlds. To address low precision and low recall, we propose a neural correction model and use it to correct false entity labels. 
We first use abstracts from Wikipedia and a well-built knowledge base, DBpedia, to annotate anchored strings in the abstracts, in order to obtain initial labels. Then, utilizing curriculum learning, we train the correction model 
and use it to correct the initial labels and train various models with the corrected dataset called AnchorNER. In addition, in order to tackle the low RAT problem of previous large but low-quality datasets, we propose a multi-task learning method for BERT models to exploit the context information. Experimental results demonstrate that our methods obtain state-of-the-art open-domain NER performances and we will release our trained BERT models as an NER toolkit called OpenNER.

Our main contributions are four-fold:
\begin{itemize}[leftmargin=*]
    \item We introduce a semi-supervised neural correction model with curriculum learning to correct the false entity labels caused by distant supervision.
    \item We use the correction model with Wikipedia and DBpedia to build a large-scale and high-quality NER dataset called AnchorNER.
    \item We propose a multi-task method to address the low RAT problem of previous open-domain NER datasets.
    \item We achieve new state-of-the-art open-domain NER results by training BERT models on our AnchorNER data.
\end{itemize}

\section{Related Work}

\noindent \textbf{Neural Networks for Named Entity Recognition}\,\,
Recently, neural networks are becoming more and more popular and they have significantly improved the performances of NER. One line of research is LSTM-based models with character-level and word-level embeddings. Some studies exploit character-level architectures \cite{DBLP:conf/coling/KuruCY16, DBLP:conf/naacl/GillickBVS16}. 
Many models also combine word-level and character-level embeddings \cite{DBLP:conf/acl/MaH16, DBLP:journals/tacl/ChiuN16, DBLP:conf/naacl/LampleBSKD16, DBLP:conf/starsem/YadavSB18}.

Another line is unsupervised pre-training methods. \cite{DBLP:conf/naacl/PetersNIGCLZ18} introduces ELMo and in \cite{DBLP:conf/coling/AkbikBV18}, authors propose contextual string embeddings which are produced with a pre-trained character language model.
BERT is proposed in \cite{DBLP:conf/naacl/DevlinCLT19} and it is pre-trained using the masked language model task and the next sentence prediction task. 
\cite{DBLP:journals/corr/abs-1903-07785} presents a pre-trained bi-directional transformer model using cloze-style word reconstruction task. 
Although pre-trained language models are proved very helpful for NER, they still need to be fine-tuned with NER datasets to obtain more information regarding entities. From this point of view, large-scale and high-quality NER datasets are still very important.

\noindent \textbf{Open-domain Named Entity Recognition}\,\,
Previously, some researchers try to build datasets for open-domain NER. The most related studies to ours are WikiNER \cite{DBLP:conf/ijcnlp/GhaddarL17} and WiNER \cite{DBLP:journals/ai/NothmanRRMC13}. The authors of the former work apply a pipeline to Wikipedia and use anchored strings and the out-link structure. The authors of the latter paper tend to classify Wikipedia articles into named entity types and link them to anchored strings. However, both datasets suffer from low precision, recall, and ratio of annotated tokens thus result in lower performances. In contrast, we implement a neural correction model and a multi-task learning method to address these problems and achieve much better results. Some other researchers explore various NER models in open-domain settings. \cite{DBLP:conf/aclnews/EkbalSFP10} studies fine-grained named entity recognition and classification in an open-domain setting but only concentrate on classes referring to people. In \cite{DBLP:conf/ijcnlp/EiseltF13}, researchers propose a two-step named entity recognizer for open-domain search queries and \cite{DBLP:conf/lrec/BowdenWOMW18} establishes a pipeline for NER in open-domain dialogue systems. Nonetheless, they still constrain their settings while our OpenNER, as an NER toolkit, can annotate texts of any genres. Besides, there are some NER toolkits, such as StanfordNLP \cite{DBLP:conf/acl/ManningSBFBM14}, spaCy\footnote{https://github.com/explosion/spaCy.}, and NLTK \cite{DBLP:journals/lre/Wagner10}. However, due to the lack of specific designs and strategies, their performances are limited in the open-domain setting.

\noindent \textbf{Noise Handling for Automatic NER Annotations}\,\,
Automatic annotations may suffer from noise and incompleteness and some methods have been proposed to solve these issues. \cite{DBLP:journals/corr/abs-1807-00745} proposes a de-noising
layer and train it on a mixture of clean and noisy data. In \cite{DBLP:conf/naacl/PaulSHK19}, people implement a similar network that trains on clean and noisy self-labeled data jointly by explicitly modeling clean and noisy labels separately. The model in \cite{DBLP:journals/corr/abs-1910-06061} clusters the training data and then compute different confusion matrices for each cluster. In \cite{DBLP:conf/eacl/SchutzeYA17}, authors introduce multi-instance multi-label learning algorithms for de-noising from distant supervision and apply them for the CF-FIGMENT task.
In \cite{DBLP:conf/coling/YangCLHZ18}, researchers utilize partial annotation learning and an instance selector to solve incomplete and noisy annotations of distant supervision for Chinese NER. 
However, these methods usually use noisy labels as training data but do not consider the information from them when testing. In this work, we design a simpler but more effective neural correction model to utilize noisy labels in both the training and testing.

\noindent \textbf{Multi-task Learning for NER}\,\,
Many previous studies explore multi-task learning for sequence labeling. Specifically, for NER, LSTM-based models are widely used \cite{DBLP:conf/naacl/LuBL19,DBLP:conf/acl/AugensteinS17}. The most related work to ours is \cite{DBLP:conf/acl/Rei17}, which analyzes that in many sequence labeling tasks, such as NER, the relevant labels are very sparse in the dataset. The issue is similar to the low RAT problem mentioned in this paper. The author proposes a multi-task learning method to predict the surrounding words for every word and it yields consistent performance improvements. Inspired by this work, we directly exploit the labels of the surrounding words for every word and our method obtains much better results.

\section{AnchorNER}
\label{sec:dataset}
To build a large-scale and high-quality NER dataset, we annotate abstracts\footnote{The first paragraph of each page.} of Wikipedia in two steps: (1) Generating distantly supervised annotations as initial labels with DBpedia. (2) Correcting initial labels with a neural correction model which will be introduced in detail in Section \ref{sec:correctionmodel}.

For the first step, we apply the following pipeline to a dump of abstracts of English Wikipedia from 2017 and obtain initial labels: (1) Extracting anchored strings in the abstract of an article. (2) Searching them in DBpedia and mapping them to their entity types mentioned in DBpedia. (3) Making exact matches in the original text using the entities and their entity types we find in DBpedia.

We use 5.2M abstracts of Wikipedia articles, consisting of 268M tokens and 12M sentences. Our motivation to utilize abstracts instead of full texts bases on the observation that entities in abstracts are denser. The tagging scheme used is the IOB2 format. 
After this step, we get an initial version of AnchorNER. The ratio of annotated tokens of the dataset is 10.22\%, which is lower than that of the manually annotated CoNLL-2003 dataset which is 16.64\%. The main reason for our relatively low ratio of annotated tokens is that we may miss the entity which does not appear in any anchored strings. The missing entities, for example, \textit{Middlesex} (LOC), \textit{British Comedy Award} (MISC), will be captured by our correction model mentioned in the next section. 

After the second step where false entity labels are corrected by the correction model, the ratio of annotated tokens of AnchorNER increases to 22.26\%. We further assess the annotation quality with a random subset of 1000 tokens and we measure the accuracy of 98\% for labels, which is better than the 92\% accuracy of WiNER. 

\section{Correction Model}
\label{sec:correctionmodel}

\subsection{Problem Statement}
\label{subsec:taskdef}


Many previous models that aim to tackle noisy labels, including false positive and false negative labels, only utilize noisy labels during training and try to predict clean labels directly during testing. However, as not all labels in the noisy data are incorrect, there is still much information we can exploit from noisy labels. Therefore, we use noisy labels as input in both training and testing processes and aim to boost the ability of our correction model to address the noise problem.

Formally, we define a sentence as a sequence of tokens $s=<w_1,w_2,\dots,w_n>$. After the first step described in Section \ref{sec:dataset}, we obtain a sequence of initial labels $\mathbf{l}$=\{$l^1$, $l^2$, ..., $l^n$\}. We input the sequence of tokens $s$ and labels $\mathbf{l}$ to the correction model and it outputs the corrected sequence of labels $\mathbf{l'}$=\{${l^1}'$, ${l^2}'$, ..., ${l^n}'$\}.


\subsection{Correction Dataset}
\label{subsec:correctiondataset}
In order to implement the correction model, we first build a correction dataset with DocRED \cite{DBLP:conf/acl/YaoYLHLLLHZS19} and the initial labels of AnchorNER. DocRED is the largest human-annotated dataset constructed from Wikipedia and Wikidata with named entities annotated. Since both DocRED and AnchorNER are obtained from Wikipedia, we find that there are 2,937 articles consisting of 8,882 sentences that appear in both datasets. Though some of the articles are not the same due to different versions, most of the entities annotated in DocRED also appear in abstracts from AnchorNER. We believe that the correction model can learn from the transformation from the initial labels in AnchorNER to the manually annotated labels in DocRED and learn a pattern of manual annotations to correct the false labels. To get the ground truth for each token in the articles which appear in both datasets, we obtain all the entities annotated in DocRED and use them to make exact matches in abstracts from AncherNER, in a descending order of the lengths of entities. If an entity matches a phrase that has not been matched, we will consider the entity type annotated in DocRED as the ground truth of the phrase. For those sentences that do not obtain any ground truth, we do not include them into the correction dataset. 
As a result, the correction dataset comprises 4,288 sentences with 121,627 tokens. Each word is assigned with two labels. The first is the initial label in AnchorNER, and the second is the ground truth obtained from DocRED.

\subsection{Model Architecture}
\label{subsec:correctionmodel}
\begin{figure}[ht!]
    \centering
    \includegraphics[width=0.6\linewidth]{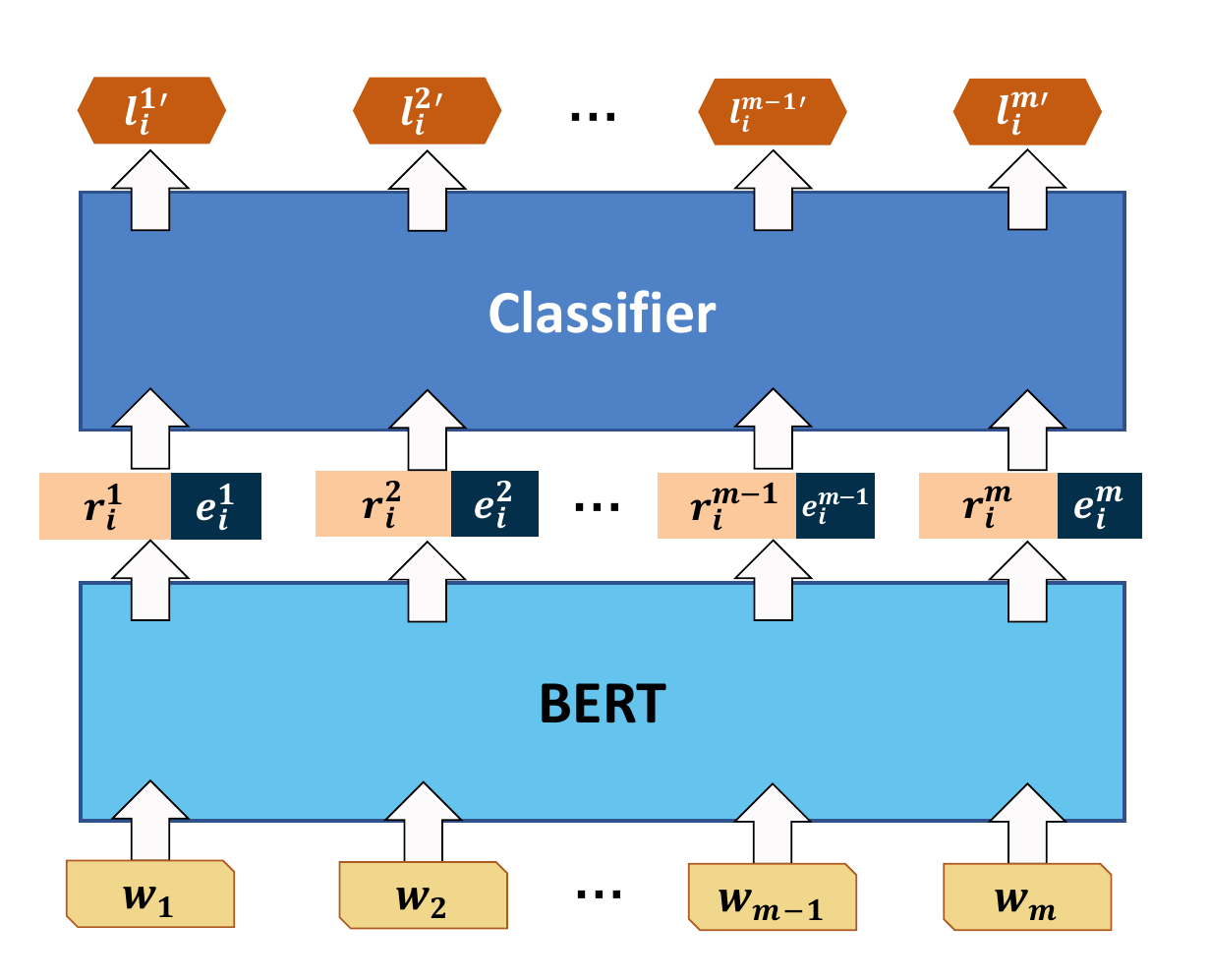}
    \caption{Overall architecture of the correction model.}
    \label{fig:correctionmodel}
\end{figure}

The architecture of the correction model is illustrated in Figure \ref{fig:correctionmodel}. We denote our correction dataset as $\mathcal{D}$=\{$\mathcal{S}$, $\mathcal{L}$, $\mathcal{L'}$ \} where $\mathcal{S}$ is the sentences in the correction dataset, $\mathcal{L}$ is the initial labels from AnchorNER and $\mathcal{L'}$ is the entity labels from DocRED. We define $\mathcal{S}$=\{$\mathbf{s_1}$, $\mathbf{s_2}$, ..., $\mathbf{s_n}$\} where $n$ is the number of sentences. $\mathbf{s_i}$=\{$w_i^1$, $w_i^2$, ..., $w_i^m$\} is a sentence where $m$ is the length of the sentence and $w_i^j$ is a word. For entity labels, $\mathcal{L}$=\{$\mathbf{l_1}$, $\mathbf{l_2}$, ..., $\mathbf{l_n}$\} and $\mathcal{L'}$=\{$\mathbf{l_1'}$, $\mathbf{l_2'}$, ..., $\mathbf{l_n'}$\} where $\mathbf{l_i}$=\{$l_i^1$, $l_i^2$, ..., $l_i^m$\} and $\mathbf{l_i'}$=\{${l_i^1}'$, ${l_i^2}'$, ..., ${l_i^m}'$\}.

For a sentence $s_i\in$\textit{$S$}, we first input it into a BERT model to obtain its representations $\mathbf{r_i}=\{{r_i^1}, {r_i^2}, ..., {r_i^m}\}$. Before the last classification layer, we embed the initial entity labels $\mathbf{l_i}$ from AnchorNER and concatenate the entity label embeddings and the sentence representations in the corresponding positions. We denote $\mathbf{e_i}=Embed(\mathbf{l_i})$ and $\mathbf{c_i}=\mathbf{e_i}\bigoplus \mathbf{r_i}$ where $\bigoplus$ means concatenation. Therefore, the last classification layer can be defined as $CLS(\mathbf{c_i})=p(\mathbf{l_i'}|\mathbf{c_i};\theta)$ where $\theta$ denotes the parameters of the classification layer. The objective function for our correction model can be defined as  
\begin{equation}
\mathcal{J}=-\sum_i p(\mathbf{l_i'}|\mathbf{c_i};\theta)
\end{equation}




\subsection{Curriculum Learning}
Curriculum learning \cite{DBLP:conf/icml/BengioLCW09} is a training strategy proposed in \cite{DBLP:conf/icml/BengioLCW09} in the context of machine learning. They demonstrate that models can be better trained when the inputs are not randomly presented but organized in a meaningful order, such as from easy to hard. Inspired by this thought, we rank all sentences in the correction dataset from easy to hard and split the correction dataset into three sets which are input into the correction model in order. Specifically, we calculate an $F_1$ score $f_i$ for each sentence in the correction dataset with the corresponding sequences of entity label $\mathbf{l_i}$ and $\mathbf{l_i'}$.
Then, we rank all the sentences in the correction dataset according to their $F_1$ score from high to low and split the correction dataset into three sets $\mathcal{D}_1$, $\mathcal{D}_2$, and $\mathcal{D}_3$. That means in $\mathcal{D}_1$, a sentence has more similar labels $\mathbf{l_i}$ and $\mathbf{l_i'}$ and $\mathcal{D}_1$ is easier for the correction model to learn. Similarly, $\mathcal{D}_3$ is more difficult for the model to learn. We input the three sets from $\mathcal{D}_1$ to $\mathcal{D}_3$ and train our correction model with each set respectively. 

\section{Multi-task Learning Method}

According to our observations, the ratios of annotated tokens of previous open-domain NER datasets are pretty low. 
The low ratio of annotated tokens of a dataset means that there are many ``O" labels and most of the words contribute little to the training process \cite{DBLP:conf/acl/Rei17}. Meanwhile, it may deteriorate the pre-trained BERT models as various words with different representations are assigned with the same label ``O". Inspired by the method proposed in \cite{DBLP:conf/acl/Rei17}, we try to exploit the entity labels of the surrounding words for each word directly and propose a novel multi-task learning method.

In order to define the range of the context we use, we set $p$ windows for binary classifications and $q$ windows for multi-class classifications. 
Specifically, we denote a sentence as $\mathbf{s}$=\{$w_1$, $w_2$, ..., $w_m$\} where $m$ is the length of the sentence. We denote the size of the $j^{th}$ window for binary classifications as $win_j^b$ ($1 \leq j \leq p$) and multi-class classifications as $win_j^m$ ($1 \leq j \leq q$). The entity types are \{$t_1, t_2, \dots, t_k$\} where $k$ is the number of entity types.
For the $j^{th}$ window, we only focus on the context information on both sides of each word $w_i$ in the sentence, namely $w_{i-win_j:i-1}$ and $w_{i+1:i+win_j}$.  Our tasks can be categorized into three classes: (1) For a window for binary classifications, setting $k$ tasks to predict whether there is any label $t_i$ $(1\leq i \leq k)$ respectively on the left side and $k$ tasks for the right side. (2) For the $j^{th}$ window for multi-class classifications, setting $win_j^m$ tasks to predict the entity label of each word on the left side and $win_j^m$ tasks for the right side. (3) Setting a task to predict the entity label of the current word. Therefore, there are $ T = p*2*k + \sum_{j=1}^q 2*win_j^m + 1$ tasks where $p*2*k$ tasks are binary classifications and $\sum_{j=1}^q 2*win_j^m + 1$ tasks are multi-class classifications. We denote the labels of the sentence as $\mathbf{l}$=\{$\mathbf{l_1}$, $\mathbf{l_2}$, ..., $\mathbf{l_m}$\} where $\mathbf{l_i}$=\{$l_i^1$, $l_i^2$, ..., $l_i^{T}$\}. We first input the sentence into a BERT model to obtain its representations $\mathbf{r}=\{{r_1}, {r_2}, ..., {r_m}\}$ and input them into the last classification layer containing T classifiers.
We define the objective function as
\begin{equation}
\mathcal{J}_{multi}=-\sum_{i=1}^m \sum_{h=1}^{T} \alpha_h \cdot p(l_i^h|r_i;\phi_h)
\end{equation}
where $\phi_h$ denotes the parameters of the $h^{th}$ classifier and $\alpha_h$ is the weight of the $h^{th}$ task.

\section{Experiments}

In this section, we evaluate the effectiveness of the proposed methods. (1) In order to evaluate the effectiveness of our correction model, we compare it with the latest and most related method proposed in \cite{DBLP:journals/corr/abs-1910-06061}. We choose the model, K-Means-CM-Freq-IP, as it obtains the best performances in the paper. As the low-resource setting in the paper is not comparable to ours, we do not follow it to sample the training data but use the whole correction dataset. We train the model and our correction model with our correction dataset respectively and use the trained models to correct the initial labels of AnchorNER. Then, we sample 1020 tokens to make evaluations. (2) To compare our dataset AnchorNER with previous open-domain NER datasets,
we train the five experimental models with AnchorNER and WiNER respectively and evaluate them on five NER datasets. (3) To evaluate our multi-task learning method, we compare it with the method proposed in \cite{DBLP:conf/acl/Rei17} which is the most related work to ours. However, as the method is implemented with LSTM which is not comparable to BERT, we change the LSTM model to a BERT-base model in order to compare the multi-task methods only. We train them with WikiNER as it has the lowest ratio of annotated tokens.

\subsection{Experiment Setup}
\paragraph{Experimental Models}
$\\$
    \textbf{Bi-LSTM-CRF} \cite{DBLP:journals/corr/HuangXY15} A bidirectional Long Short-Term Memory (LSTM) network with a Conditional Random Field (CRF) layer.
    
    \noindent \textbf{CVT} \cite{DBLP:conf/emnlp/ClarkLML18} A semi-supervised learning algorithm that improves the representations of a Bi-LSTM sentence encoder using a mix of labeled and unlabeled data.
   
    \noindent \textbf{ELMo} \cite{DBLP:conf/naacl/PetersNIGCLZ18} A type of deep contextualized word representation which models both complex characteristics of word use and how these uses vary across linguistic contexts.
    
    \noindent \textbf{Flair} \cite{DBLP:conf/naacl/AkbikBV19,DBLP:conf/coling/AkbikBV18} A type of contextual word representation which is distilled from all contextualized instances using a pooling operation.
    
    \noindent \textbf{BERT} \cite{DBLP:conf/naacl/DevlinCLT19} A language representation model which is designed to pre-train deep bidirectional representations from the unlabeled text by jointly conditioning on both left and right context in all layers.

\paragraph{Datasets}
$\\$
    \noindent\textbf{CoNLL03} \cite{DBLP:conf/conll/SangM03} The CoNLL 2003 Shared Task dataset is a well known NER dataset built up with Reuters newswire articles. It is annotated with four entity types (PER, LOC, ORG, and MISC). 
    
    \noindent\textbf{Ontonote5} \cite{DBLP:conf/conll/PradhanMXUZ12} The OntoNote 5.0 dataset contains newswire, magazine articles, broadcast news, broadcast conversations, web data, and conversational speech data. The dataset has about 1.6M words and is annotated with 18 named entity types. We follow \cite{nothman2008learning} to map annotations to CoNLL 2003 tag set.
    
    \noindent\textbf{Tweet} \cite{DBLP:conf/emnlp/RitterCME11} Authors annotate 2,400 tweets (34k tokens) with 10 entity types in \cite{DBLP:conf/emnlp/RitterCME11} and we map the entity types to CoNLL 2003 tag set.
    
    \noindent\textbf{Webpage} \cite{DBLP:conf/conll/RatinovR09} Researchers manually annotated a collection of 20 webpages (8k tokens) on different topics with the CoNLL 2003 NE classes in \cite{DBLP:conf/conll/RatinovR09}.
    
    \noindent\textbf{WikiGold} \cite{DBLP:conf/acl-pwnlp/BalasuriyaRNMC09} Researchers manually annotate a set of Wikipedia articles comprising 40k tokens with the CoNLL 2003 tag set in \cite{DBLP:conf/acl-pwnlp/BalasuriyaRNMC09}.
\paragraph{Evaluation Metrics}
For the comparison between our correction model and K-Means-CM-Freq-IP, we evaluate accuracy (Acc), precision, recall, and F1-score. For the comparison between AnchorNER and WiNER and the comparison between multi-task methods, we use span-level F1-score for four classes (LOC, PER, ORG, and MISC)\footnote{Note that in WiNER paper, authors use token-level F1-score for three classes (LOC, PER, and ORG). Therefore, our results are not comparable to theirs.}.





\begin{table}[t]
\setlength{\abovecaptionskip}{0.3cm}
\setlength{\belowcaptionskip}{0cm}
\small
\centering
\scalebox{1.1}{%
\begin{tabular}{lcccc}
     \toprule
     \bf Model & \bf Acc & \bf Precision & \bf Recall & \bf F1\\
     \midrule
     K-Means-CM-Freq-IP & 94.80 & 63.21 & 67.90 & 65.47\\
     Correction Model& \bf 98.13 & \bf 85.06 & \bf 90.12 & \bf 87.52\\
     \bottomrule
\end{tabular}%
}
\caption{Comparing our correction model with K-Means-CM-Freq-IP proposed in \cite{DBLP:journals/corr/abs-1910-06061}.}
\label{tab:noise model}
\end{table}

\begin{table*}[t]
\small
\begin{center}
\begin{tabular}{llcccccc}
     \toprule
     \bf Base Model&\bf Method & \bf CoNLL03 &\bf Ontonote5 &\bf Tweet &\bf Webpage &\bf Wikigold & \textbf{Avg.} \\
     \midrule 
     \multirow{3}{*}{\emph{Public NER toolkits}} &
     NLTK & 48.91   & 39.00 & 21.18 & 28.17 & 44.22 & 36.30 \\
     &SpaCy & 65.14   & \textbf{76.66} & 31.87 & 36.39 & 58.86 & 53.78 \\
     &StanfordNER & 87.95   & 60.64 & 35.74 & 44.34 & 62.00  & 58.13 \\ 
     \midrule
     \multirow{3}{*}{\emph{Bi-LSTM-CRF}} &
     WiNER & 85.83 & 58.53 & 36.60 & 51.40 & 66.67  & 59.81\\
     &Ours& 86.36 & 58.46 & 38.32 & 52.84 & 68.38  & 60.87(+1.78\%)\\
     \midrule
     \multirow{3}{*}{\emph{CVT}} &
     WiNER & 92.98 & 64.77 & 41.53 & 55.77 & 76.68  & 66.35\\
     &Ours& \bf 93.23 & 66.08 & 41.79 & \bf 58.50 & 79.22  & 67.76(+2.14\%)\\
     \midrule
     \multirow{3}{*}{\emph{ELMo}} &
     WiNER & 92.72 & 65.91 & 41.03 & 57.16 & 77.99  & 66.96\\
     &Ours& 92.75 & 64.72 & 41.77 & 58.26 & 80.03  & 67.51(+0.81\%)\\
     \midrule
     \multirow{3}{*}{\emph{Flair}} &
     WiNER & 92.99 & 65.36 & 41.92 & 51.61 & 76.73  & 65.72\\
     &Ours& 93.01 & 65.39 & 40.98 & 54.36 & 77.70  & 66.29(+0.86\%)\\
     \midrule
     \multirow{3}{*}{\emph{$BERT_{base}$}} &
     WiNER & 91.60 & 66.17 & 47.16 & 48.87 & 79.91  & 66.74\\
     &Ours & 92.02 & 67.53 & 50.22 & 48.75 & \bf 82.37  & 68.18(+2.16\%)\\
     \midrule
     \multirow{3}{*}{\emph{$BERT_{large}$}} &
     WiNER & 92.66 & 68.47 & 49.54 & 49.44 & 80.36  & 68.09\\
     &Ours& 92.15 & 68.19 & \bf 51.45 & 50.81 & 82.16  & \bf 68.95(+1.26\%)\\
     \bottomrule
\end{tabular}
\end{center}
\caption{Comparing various models trained with AnchorNER and WiNER on various open-domain NER datasets.}
\label{tab:result}
\end{table*}

\begin{table}[t]
\setlength{\abovecaptionskip}{0.3cm}
\setlength{\belowcaptionskip}{0cm}
\small
\centering
\scalebox{1.1}{%
\begin{tabular}{lcccc}
     \toprule
     \bf Dataset & \bf WikiNER & \bf WiNER & \bf CoNLL & \bf AnchorNER\\
     \midrule
     RAT(\%) & 13.93 & 15.19 & 16.65 & \bf 22.26\\
     \bottomrule
\end{tabular}%
}
\caption{Ratio of annotated tokens of various NER datasets.}
\label{tab:coverage}
\end{table}

\begin{table*}[t]
\small
\centering
\begin{tabular}{lcccccc}
     \toprule 
     \bf Method & \bf CoNLL03   & \bf Ontonote5 & \bf Tweet & \bf Webpage & \bf Wikigold & \textbf{Avg.} \\
     \midrule 
     Single & \bf 91.06 & \bf 66.25 & 48.76 & 46.85 & 76.65  & 65.91 \\
     \cite{DBLP:conf/acl/Rei17} & 90.98 & 64.79 & 44.37 & 48.36 & 79.61  & 65.62 \\
     Multi-task & 91.04   & 65.61 & \bf 49.40 & \bf 50.18 & \bf 80.91 & \bf 67.43 \\
     \bottomrule
\end{tabular}%
\caption{Comparing our multi-task learning method with the method proposed in \cite{DBLP:conf/acl/Rei17}.}
\label{tab:multi}
\end{table*}

\begin{table*}[t]
\small
\begin{center}
\begin{tabular}{lcccccc}
     \toprule 
     \bf Model & \bf CoNLL03   & \bf Ontonote5 & \bf Tweet & \bf Webpage & \bf Wikigold & \textbf{Avg.} \\
     \midrule 
    $\rm OpenNER_{\rm base}$ & 92.02 & \textbf{67.53} & \textbf{50.22} & 48.75 & \textbf{82.37}  & \textbf{68.19} \\
     w/o Wiki label & 92.11   & 67.29 & 49.50 & 48.56 & 81.20 & 67.73 (-0.65\%) \\ 
     smaller correction dataset & \textbf{92.13}   & 67.29 & 48.57 & 47.09 & 81.90 & 67.40 (-1.15\%) \\
     w/o curriculum learning & 91.58   & 66.94 & 46.98 & 47.21 & 81.05& 66.75 (-2.09\%) \\
     w/o correction & 91.86   & 66.44 & 46.75 & 46.29 & 78.41 & 65.95  (-3.27\%) \\ 
     \midrule
     \multicolumn{7}{c}{\underline{\emph{Ablation on fine-tuning with labeled data}}}\\
     CoNLL (i.e.$\rm BERT_{\rm base}$) & 91.95  & 66.63 & 47.79 & 45.58 & 78.21 & 66.03(-3.15\%) \\
     DocRED & 70.25  & 56.82 & 47.03 & \bf 50.68 & 76.76 & 60.31 (-11.54\%) \\
     DocRED+CoNLL~(mixed) & 90.63  & 64.89 & 43.25 & 49.30 & 77.44 & 65.10 (-4.51\%) \\ 
     DocRED+CoNLL~(sequential) & 91.69  & 66.24 & 47.86 & 48.23 & 79.79 & 66.76 (-2.08\%) \\ 
     
     \bottomrule
\end{tabular}
\end{center}
\caption{Ablation Study Results.}
\label{tab:ablation}
\end{table*}

\paragraph{Implementation Details}



We first split AnchorNER into 8:1:1 for  training, validation, and testing according to the categories of different titles. Due to the limitation of computing resources, we randomly sample 70,000, 10,000, and 10,000 abstracts for training, validation, and testing respectively and they totally contain three million tokens. 
We use the cased BERT model and the maximum sequence length is 128. During training, we set the batch size as 32 for BERT-base and 8 for BERT-large. The optimizer is BERTAdam \cite{DBLP:conf/naacl/DevlinCLT19} and the initial learning rate is 5e-5. Warming-up proportion is 0.4. In the correction model, we utilize  12-dimension one-hot vectors for embeddings of entity labels from AnchorNER. For experimental models, we use default settings and following \cite{DBLP:conf/ijcnlp/GhaddarL17}, we train the models with both CoNLL03 and WiNER/AnchorNER\footnote{We use ``train" here but for pre-training models, including ELMo, Flair and BERT, we actually fine-tune the models.}. We train the models with AnchorNER or WiNER first and then with CoNLL03. For all models, we fine-tune them with AnchorNER or WiNER for five epochs and then fine-tune them with the CoNLL03 dataset up to 50 epochs with early stopping if the performance on the development set does not improve within the last five epochs. For the multi-task models, we fine-tune BERT models with WikiNER in the multi-task setting and CoNLL03 in the normal (single-task) setting. Other settings are the same as the BERT models in experimental models. As WikiNER uses the IOB format, we convert it to the IOB2 format. Based on the performance on the development set, we set $p=2$ and window sizes as 2 and 4 respectively. We set $q=1$ and a window size of 1. We use equal weights for all tasks and $\alpha_h=\frac{1}{T}$ ($1 \leq h \leq T$).

\subsection{Results}
\label{subsec:mainresults}
\paragraph{Correction Model}
The results are shown in Table \ref{tab:noise model}. We find that our correction model outperforms K-Means-CM-Freq-IP in all metrics. The reason why K-Means-CM-Freq-IP obtains low performances may be that the IOB2 format would increase matrix size and make the confusion matrix estimation more difficult \cite{DBLP:journals/corr/abs-1910-06061}. In contrast, our model does not have the restriction and is strong enough to learn the pattern of manual annotations from small but clean datasets.

\paragraph{AnchorNER}
The performance of each model is presented in Table \ref{tab:result}. From the results, we have the following observations: (1) Various models trained with AnchorNER outperform those trained with WiNER. (2) The BERT models trained with AnchorNER are comparable to the state-of-the-art results on the benchmark dataset CoNLL03. Meanwhile, AnchorNER obtains the highest ratio of annotated tokens as shown in Table \ref{tab:coverage}.

\paragraph{OpenNER}
Our released open-domain NER toolkit, OpenNER, includes the BERT-base model ($\rm OpenNER_{\rm base}$) and the BERT-large model ($\rm OpenNER_{\rm large}$) trained with AnchorNER. As shown in Table \ref{tab:result}, our toolkit OpenNER significantly outperforms existing NER toolkits on five NER datasets.

\paragraph{Multi-task}
The results are presented in Table \ref{tab:multi}. Compared with the single-task (normal) method, our multi-task method performs better on three datasets and the average score. In \cite{DBLP:conf/acl/Rei17}, authors utilize a language modeling objective on which different texts have different distributions, especially for tweets. This may be the reason why the method obtains some poor performances.

\subsection{Ablation Studies}
Now, in order to analyze the effectiveness of each component of our method, we focus on the BERT-base model and conduct some ablation studies to systematically look into some important components of our method.

\paragraph{Effectiveness of AnchorNER}
We evaluate the effectiveness of AnchorNER by not using AnchorNER. Instead, we train a BERT-base model directly with the DocRED and CoNLL dataset. We try two training methods: (1) First using the DocRED dataset and then using the CoNLL dataset to fine-tune the BERT model twice. (2) Using the mix of the DocRED dataset and the CoNLL dataset to fine-tune the BERT model. These two methods reduce the performance by 2.08\% and 4.51\% respectively as shown in Table \ref{tab:ablation}.

\paragraph{Effectiveness of Initial Labels of AnchorNER}
We remove the initial labels of AnchorNER from the correction model to evaluate its effectiveness. In this way, the correction model becomes an original BERT classifier trained with DocRED. Results show that removing these labels slightly hurts the performance of the model, reducing the average score by 0.65\%.

\paragraph{Effectiveness of the Size of the Correction Dataset} 
We try to restrict the collection of the correction dataset, leaving only the exact same sentences that appear in both AnchorNER and DocRED. A total of 2,587 sentences consisting of 61,343 tokens meets this requirement, which is almost half the size of our correction dataset. The results in the third row of Table \ref{tab:ablation} illustrate that the overall performance using this smaller correction dataset decreases by 1.15\%.

\paragraph{Effectiveness of Correction}
We study the effectiveness of correction by removing the process of correction and only using the initial labels of AnchorNER. As we mentioned in Section \ref{sec:dataset}, we will miss some entities in this way, resulting in a lower recall rate. The results in the fifth row of Table \ref{tab:ablation} show that removing the correction part leads to a reduction of 3.27\%  in average score.

\paragraph{Effectiveness of Curriculum Learning} 

During the training of the correction model, we take advantage of the idea of curriculum learning. We compare this training method with a variant, in which we directly train the correction model with the whole correction dataset. From Table \ref{tab:ablation} we can see that if we do not use curriculum learning, the performance decreases by 2.09\%.

\subsection{Qualitative Study of the Correction Model}
\paragraph{Corrected v.s. uncorrected AnchorNER}
We compare the labels of the same 1000 tokens in corrected and uncorrected AnchorNER. We find that 61 entities are retrieved after correction.

\begin{figure}[!ht]
\subfigure[A case that the correction model retrieves the missing entity called \it British Comedy Award.]{\centering
    \includegraphics[width=1.0\linewidth]{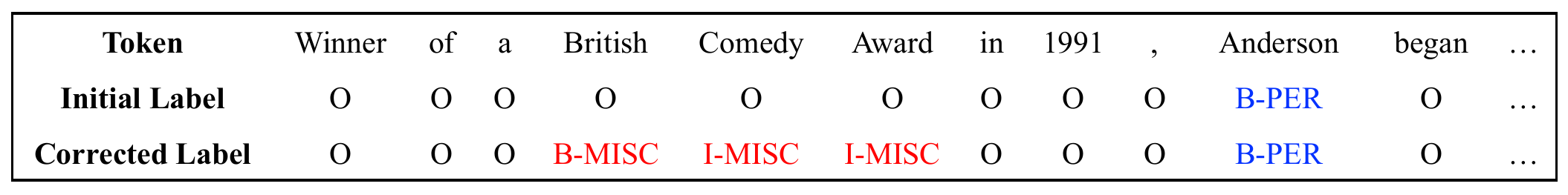}
    \label{fig:correctexample1}
}
\subfigure[A case that the correction model retrieves two missing entities called \textit{CANDU} and \textit{Canada Deuterium Uranium}.]{\centering
    \includegraphics[width=1.0\linewidth]{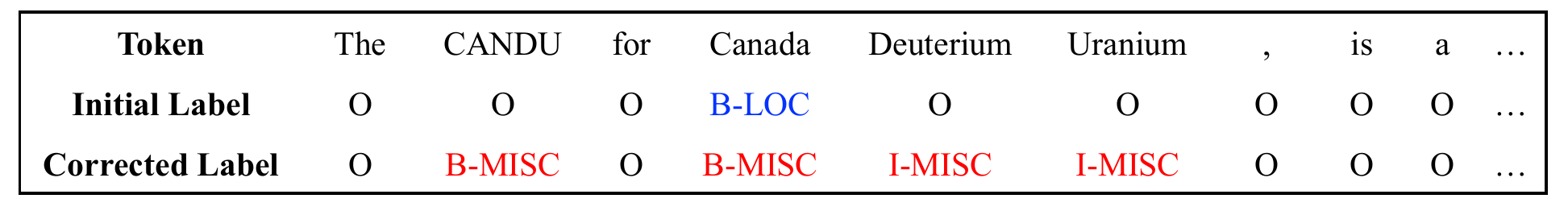}
    \label{fig:correctexample2}
}
\subfigure[A case that the correction model fails to recognize the entity called \textit{Two Flint}.]{\centering
    \includegraphics[width=1.0\linewidth]{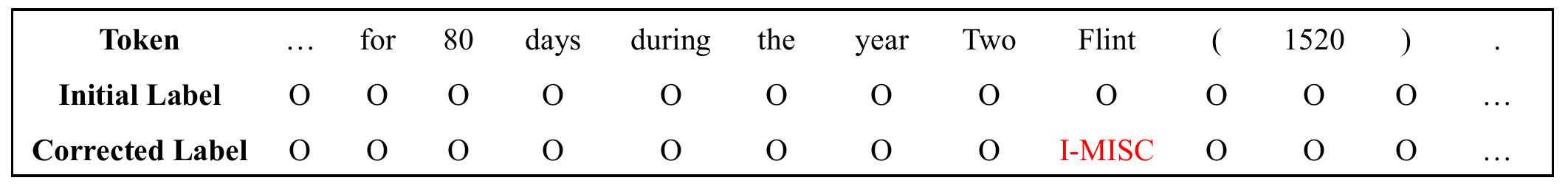}
    \label{fig:correctexample3}
}

    \caption{Comparison between the dataset before correction and the dataset after correction.}
    \label{fig:correctexample}
    
\end{figure}

\paragraph{Examples}
Figure \ref{fig:correctexample} shows two examples of correction. In the sentence in Figure \ref{fig:correctexample1}, the entity \textit{British Comedy Award} is not recognized during the three-step process mentioned in Section \ref{sec:dataset} because it does not appear in any hyperlinks. The second example in Figure \ref{fig:correctexample2} shows that the entities \textit{CANDU} and \textit{Canada Deuterium Uranium} are not recognized. In the uncorrected dataset, \textit{Canada} is mislabeled as B-LOC. This is because \textit{Canada} appears in one of the anchored strings, but \textit{Canada Deuterium Uranium} does not. With the help of the correction model, all the labels are corrected.

However, we also identify some errors during the correction process as shown in Figure \ref{fig:correctexample3}. \textit{Two Flint} should be marked as MISC, but the correction model fails to recognize it. Instead, it only tags \textit{Flint} as I-MISC. This kind of error occurs because the model predicts the label for each token independently and fails to recognize the whole entity. Another type of error is that the model modifies the correct label to the wrong label. 

\subsection{Discussion}
In order to address the low qualities of the open-domain NER datasets built with distant supervision, we propose methods from two aspects: (1) Introducing external supervision to correct the false labels. (2) Exploiting the internal context information. 
The correction model
, trained with the small but high-quality DocRED, 
is strong external supervision.
Using it, we build AnchorNER which covers various entities and achieve better open-domain NER performances. 
Our multi-task learning method makes ``O" labels different from each other based on their context.
Previous open-domain NER datasets
suffer from the low RAT problem and our method is proposed to solve it. However, built with our correction model, AnchorNER does not have such problem anymore.

\section{Conclusion and Future Work}
In this paper, we propose a neural correction model to correct false labels caused by distant supervision. We use it with abstracts from Wikipedia and DBpedia to obtain a large and high-quality dataset called AnchorNER. To address the low RAT problem of previous open-domain NER datasets, we introduce a multi-task learning method to exploit the context information. Our correction model, AnchorNER and multi-task learning method obtain better performances than corresponding baselines. 
In the future, we would like to utilize AnchorNER during the pre-training process and come up with more pre-training methods leveraging the information about entities.

\bibliographystyle{unsrt}
\bibliography{references}  






\end{document}